\documentclass[11pt]{article}

\usepackage[final]{acl}

\usepackage{times}
\usepackage{latexsym}

\usepackage[T1]{fontenc}

\usepackage[utf8]{inputenc}

\usepackage{microtype}
\usepackage[most]{tcolorbox}

\usepackage{inconsolata}

\usepackage{graphicx}

\usepackage{amsmath}
\usepackage{amssymb}
\usepackage{booktabs}
\usepackage{multirow}

\newcommand{\conf}{\texttt{<conf>}}
\newcommand{\ci}[1]{{\scriptsize\mbox{[#1]}}}
\newcommand{\cit}[1]{{\fontsize{7}{8}\selectfont\mbox{[#1]}}}
\newcommand{\x}{$\times$}

\title{Can an Agent Learn to Predict Its Own Failure?\\Outcome-Grounded Confidence Calibration in LLM Agents}

\author{
    Ashutosh Hathidara \\
    SAP \\
    {\small \texttt{ashutosh.hathidara@sap.com}}
}

\begin{document}
\maketitle
\begin{abstract}
Controlling an acting LLM agent means deciding, at each step of a long-horizon task, whether to trust its next action, abstain, or hand off to a human. That decision needs a per-step confidence signal predicting whether the current rollout will \emph{eventually} succeed. The agent's own self-reported verbalized confidence is the natural candidate, but existing reinforcement learning methods calibrate it against the correctness of a decision at the current step. We propose a method to calibrate it against the eventual outcome of a multi-step episode: under GRPO, the group of rollouts for a task supplies success labels, and a proper scoring rule scores each step's confidence against the realized outcome. In offline RL on recorded trajectories of a Qwen2.5-7B agent in ALFWorld, TextWorld, and WebShop environments, the AUROC of step confidence against episode outcome rises from near chance to \textbf{0.68--0.85}, and the \textbf{top-10\%} of steps the agent is most confident about succeed \textbf{$\sim$4.3$\times$} more often than average. In online RL on the live ALFWorld agent, four calibration rewards improve early-step AUROC, expected calibration error goes from \textbf{0.42} to below \textbf{0.12}, and selective execution improves over both the untrained agent and a format-only RL control, without degrading task success. We also introduce an evaluation protocol that separates episode depth, task difficulty, and attempt-level foresight, which shows where the gain lies: the trained agent recognizes from its first step which tasks it will fail, while telling a succeeding attempt from a failing one at the same task is harder, with some early-window signal emerging on TextWorld.
\end{abstract}

\section{Introduction}

Acting LLM agents increasingly operate in long-horizon, partially observable environments \citep{Shridhar2021,Yao2022,Cote2018,Shridhar2020,Hausknecht2020,Wang2022,Zhou2024,Liu2024}, where a single episode spans dozens of actions under compounding uncertainty. \emph{Controlling} such an agent at runtime requires an estimate of the probability that the current attempt will succeed, so that it can continue with the action, verify with observation, or escalate to a human~\citep{Kadavath2022}. Such a probability estimate can work as a confidence signal that predicts whether the current rollout will ultimately reach the goal, not whether the current step is locally correct.

The natural candidate is a self-reported verbalized confidence of an agent. Without additional focused training, smaller LLMs are poorly calibrated and barely predictive (\S\ref{sec:results}). Previous works on Reinforcement Learning (RL) for calibration \citep{Damani2025,BaniHarouni2025,Wang2026b,Xuan2026,Zhang2026} focus on training confidence to predict either a single answer or step-level correctness. Such greedy optimization signals train the model to be locally predictive, promising no predictiveness for eventual outcome.

We propose a methodology to train a step-wise confidence signal that could predict the eventual outcome of the task. Under group-relative policy optimization (GRPO), the rollouts for a task supply success labels at no extra cost. We define the per-step reward as a proper scoring rule on the step confidence against the realized outcome, with other reward terms. We conduct offline \& online RL training. Our evaluation protocol separates what the aggregate score mixes (\S\ref{sec:estimands}).

Our approach reveals that outcome-grounded calibration can turn a weak agent's confidence into a reliable early signal of eventual outcome, consistently across environments and training regimes. Our evaluation protocol then reveals precisely where this new ability comes from, and where it does not.

In summary, our contributions include: (i) Outcome-grounded calibration training under GRPO, (ii) an evaluation protocol for episodic confidence, and (iii) an empirical analysis of what an LLM learns during such a training process.


\section{Related Work}
\label{sec:related}

\paragraph{Measuring and eliciting confidence.}
A broad line of work reads a confidence out of a model without additional training: verbalized probabilities \citep{Lin2022,Tian2023}, self-consistency and self-evaluation \citep{Kadavath2022}, and elicitation strategies for black-box models \citep{Xiong2024}. A related line of work documents that post-hoc calibration of neural classifiers degrades with scale and training \citep{Guo2017}. These works primarily target \emph{synchronous} single-answer correctness, where confidence degrades on weak models, precisely the area we target.

\paragraph{Reinforcement learning for confidence.}
The most closely related work trains calibration. RLCR \citep{Damani2025} optimizes a Brier reward for verbalized confidence, and related efforts tie confidence to reward shaping, uncertainty, or abstention in non-agentic settings \citep{BaniHarouni2025,Xu2024,Zhang2026,Leng2024,Yaldiz2026,Li2026,Wang2026}. All optimize the calibration of a \emph{single synchronous output}; none targets the eventual success of an agent through a long-horizon task.

\paragraph{Positioning against concurrent agentic methods.}
Three concurrent methods are closest. CAR/MSCR \citep{Xuan2026} and CAPO \citep{Wang2026b} train confidence in tool-use and reasoning settings, and RLCR \citep{Damani2025} in single-answer QA; all optimize confidence against the \emph{synchronous} correctness of the current answer or step, whereas we target the \emph{predictive} success of the eventual rollout. They differ from us in how the signal is produced and rewarded: CAPO reads a perplexity-derived confidence and reshapes the advantage estimator through a consistency-and-regret construction, and MSCR uses a margin reward, while we train an explicit verbalized token under a lightweight per-completion calibration reward.

\paragraph{Selective prediction.}
Our governance mechanism is an instance of selective prediction: we evaluate it with risk-coverage curves \citep{Geifman2017,Kamath2020} and train the signal with classical proper scoring rules \citep{Brier1950,GneitingRaftery2007}. Closest in spirit, belief-uncertainty agents such as \citet{Singh2026} estimate an agent's internal belief rather than its outcome-predictive confidence; we distinguish the two in \S\ref{sec:method}.


\section{Problem Setup and Evaluation Protocol}
\label{sec:prob}

\label{sec:setup}

We first state what a per-step confidence should predict (\S\ref{sec:target}), then define the family of estimands that separates genuine foresight from the depth and difficulty confounds a pooled score mixes together (\S\ref{sec:estimands}), and finally fix the calibration, selective-execution, and statistical conventions used throughout (\S\ref{sec:stats}).

\subsection{Predictive Confidence Target}
\label{sec:target}

We model an acting agent as a partially observable decision process with policy $\pi_\theta$. At step $t$ the policy observes history $h_t$ (the prompt, prior actions, and environment feedback) and emits an action together with an inline confidence, $(a_t, c_t)\sim\pi_\theta(\cdot\mid h_t)$, where $c_t\in[0,1]$ is parsed from a \texttt{<conf>NN</conf>} tag (an integer $0$--$100$ rescaled to $[0,1]$; a missing or malformed tag defaults to a neutral $0.5$). A full rollout $\tau$ terminates and receives a binary success label $Y\in\{0,1\}$ ($Y{=}1$ iff the task is solved).

We define the \textbf{predictive-calibration target}: $c_t$ should track
\begin{equation}
  c_t \;\approx\; P(Y=1 \mid h_t),
  \label{eq:target}
\end{equation}
the probability that \emph{this episode} ultimately succeeds given the history so far. We target eventual success rather than the \emph{synchronous} $P(\text{current step correct})$ used by prior work. Long-horizon agentic tasks admit many valid solution paths, so multiple actions may be correct at each step; local correctness is thus ill-defined, and only the eventual outcome matters as a label.

\subsection{Estimands}
\label{sec:estimands}

Task $s$ has attempts $i=1,\dots,G$ with outcomes $Y_{s,i}$ and step confidences $c_{s,i,t}$, every step inheriting its attempt's outcome. A step's confidence can rank correctly for three unrelated reasons: (i) the task is easy, (ii) the attempt is going well, or (iii) the episode is near its end. A single pooled metric conflates all three, so we report four \emph{estimands}, each an AUROC against episode success---the probability that a step of a succeeding attempt is higher than a step of a failing one---differing only in which steps may be compared. We evaluate over all steps, the early window ($k\le2$, steps 0--2), and step 0 alone.

\emph{Pooled AUROC} compares any success-step with any failure-step and holds none of the confounds fixed. \emph{Between-task AUROC} collapses each task to its mean confidence, leaving only task difficulty~(i). \emph{Within-task AUROC} compares succeeding against failing attempts of the same task, cancelling difficulty to isolate attempt-level foresight~(ii). \emph{Depth-matched within-task AUROC} compares them only at the same step index, so neither difficulty nor depth contributes and only foresight survives~(iii). Against each we report a \emph{depth-only model-free baseline} scoring every step by $-t$: over all steps it can match or beat a real signal, but the early window and step~0 drive it to chance, so any gain there reflects foresight rather than position. At step~0 every attempt sees the same observation, so a gain can arise only from ranking tasks or from the first sampled action, before any environment feedback. All formal definitions appear in Appendix~\ref{app:estimands}.

\subsection{Calibration \& Selective Execution}
\label{sec:stats}

We report two kinds of metrics. For \textbf{calibration}, how well confidence matches outcome frequency: Expected Calibration Error (ECE) with 10 equal-width bins \citep{Guo2017}, and the Brier score \citep{Brier1950}. For \textbf{selective execution}, how useful confidence is for gating actions: the success rate among the top-$q$\% most confident steps in a window ($q=10,25$), and task-level triage (the success rate when only the top 50\% of tasks by mean early confidence are attempted). Online intervals are 95\% task-level cluster bootstraps. Offline intervals are 95\% bootstraps over steps; steps of one trajectory share a label, so these intervals understate uncertainty. Paired deltas resample the tasks shared by both runs under comparison; within-task deltas use only the tasks that are mixed-outcome in both arms, a small subset, so these intervals rest on few clusters and are correspondingly wide. Tables~\ref{tab:onpolicy} and \ref{tab:decomp} and the appendix report $\sim$100 paired deltas.


\section{Methodology}
\label{sec:method}

Our method turns the rollouts that GRPO already samples into a training signal for confidence. The $G$ attempts at a task each end in an outcome that labels every one of their steps, so a step can be scored by how well its stated confidence matches the attempt outcome it belongs to. We assemble this into three pieces: a group-relative advantage that credits a step for both the eventual outcome and the quality of its confidence (\S\ref{sec:grpo}), a per-step reward built from a proper scoring rule that supplies the confidence term, and an optional spread term that keeps confidence away from the endpoints of the scale (\S\ref{sec:reward}). The same signal drives training on recorded offline trajectories and on the live agent alike (\S\ref{sec:routes}).

\begin{figure}[t]
  \centering
  \includegraphics[width=\linewidth]{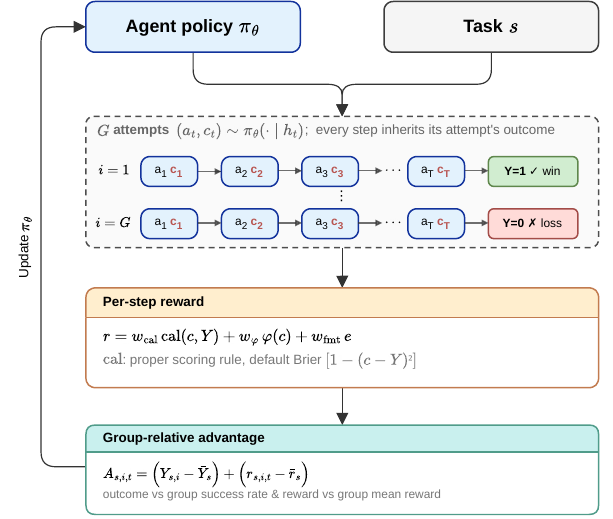}
   \caption{\textbf{Outcome-grounded calibration under GRPO.}
  The $G$ attempts at a task end in outcomes $Y$ that label every one of their steps.
  Each step's confidence $c$ is scored against that label by a proper scoring rule
  (Eq.~\ref{eq:reward}), and the group-relative advantage
  credits the step for both the outcome and the quality of its confidence before
  updating $\pi_\theta$.}
  \label{fig:method}
\end{figure}

\subsection{Group-Relative Training with Outcome}
\label{sec:grpo}

GRPO \citep{Shao2024,DeepSeekAI2025} samples $G$ rollouts of one task and forms advantages relative to the group mean. In an episodic setting the realized outcomes $Y_i$ of the group are the labels of Eq.~\eqref{eq:target} at no extra cost. Online, the advantage of every step is
\begin{equation}
A_{s,i,t} = \big(Y_{s,i}-\bar Y_s\big) + \big(r_{s,i,t}-\bar r_s\big),
\label{eq:adv}
\end{equation}
where $\bar Y_s=\tfrac{1}{G}\sum_i Y_{s,i}$ is the group's success rate, $\bar r_s$ the group mean of the per-step rewards, and $r_{s,i,t}$ is the per-step reward below. We also ablate two alternative optimization strategies against this GRPO objective (Section~\ref{sec:offline}): RLOO \citep{Ahmadian2024}, a leave-one-out variant that baselines each rollout against the mean of the others rather than the full group, and PPO with a learned value head, a non-group-relative control that replaces the group baseline with a learned critic.

\subsection{Per-Step Reward}
\label{sec:reward}

The per-step reward is
\begin{equation}
r = w_{\text{cal}}\,\mathrm{cal}(c,Y) + w_\varphi\,\varphi(c) + w_{\text{fmt}}\,e,
\label{eq:reward}
\end{equation}
where $e\in\{0,1\}$ is a format indicator that is $1$ iff the step emits a well-formed \conf{} tag (a missing or malformed tag sets $e=0$ and defaults $c$ to $0.5$; \S\ref{sec:target}), and $w_{\text{fmt}}=0.2$ throughout.
By default we instantiate the calibration term $\mathrm{cal}(\cdot)$ of Eq.~\eqref{eq:reward} with the Brier reward,
\begin{equation}
\mathrm{cal}(c,Y)=r_B(c,Y)=1-(c-Y)^2,
\label{eq:brier}
\end{equation}
which is strictly proper: with $p=P(Y=1\mid h)$, $\mathbb{E}[r_B\mid h]=1-\mathrm{Var}(Y\mid h)-(c-p)^2$, maximized at $c=p$ \citep{Brier1950,GneitingRaftery2007}. Any strictly proper rule shares this optimum, so the form should not matter; Section~\ref{sec:offline} tests this with other choices of $\mathrm{cal}(\cdot)$: the logarithmic and spherical rules (proper), L1 (improper), a decision/selective reward, and a ranking surrogate. The ranking surrogate is a streaming soft-AUROC against moving averages of opposite-class confidence; it reads only the order of confidences (all forms are in Appendix~\ref{app:hyper}). Balanced Brier reweights wins and losses within a group so that both contribute equally, with weight $0.5/\bar Y_s$ on wins and $0.5/(1-\bar Y_s)$ on losses.

The spread term $\varphi(c)=4c(1-c)$ is concave, maximal at $c=\tfrac12$, and zero at the endpoints; with weight $w_\varphi$ it is a regularizer that keeps confidence in the interior of the scale---preventing the calibration reward from collapsing it onto $0$ or $1$---at some cost in ranking quality (\S\ref{sec:onpolicy}). The dispersion variant $\varphi_{\text{disp}}(c)=\min\!\big(1,4(c-\bar c_s)^2\big)$, with $\bar c_s$ the group's mean confidence, replaces $\varphi$ and so rewards spread around the group mean rather than toward $c=\tfrac12$.

\subsection{Training Strategies}
\label{sec:routes}

We use the standard distinction between offline RL, which trains on a fixed dataset of recorded trajectories without interacting with the environment, and online RL, which trains the agent on the trajectories it generates in the live environment (on-policy).
\paragraph{Offline Training.} The offline route runs GRPO on (prefix $\to$ action + \conf{}) rows cut from one recorded untrained Qwen2.5-7B-Instruct \citep{Qwen2024} trajectory per game, each prefix labeled with that trajectory's outcome. The per-step reward is the weighted composite
\begin{equation}
r^{\text{off}} = 0.5\,a + 0.2\,e + w_{\text{cal}}\,\mathrm{cal}(c,Y) + w_\varphi\,\varphi(c),
\label{eq:reward_off}
\end{equation}
where $a\in\{0,1\}$ indicates that the sampled action matches the recorded one, $e$ and $\mathrm{cal}(c,Y)$ are the format and calibration terms of Eq.~\eqref{eq:reward}, $w_{\text{cal}}=0.3$ by default, and the spread weight $w_\varphi$ is nonzero only when regularization is enabled (Appendix~\ref{app:hyper}). The policy never acts on its own outputs. Because all eight generations for a prefix share the recorded outcome as their label, the group supplies no label contrast, the action-match term rewards reproducing the recorded action, and trained confidence is scored on recorded prefixes of the untrained policy. The reward-form arms of Table~\ref{tab:rewardform} use $w_{\text{cal}}=1.0$ in place of the 0.3 default.

\paragraph{Online Training.} The online route instead trains the live agent on its own rollouts for 120 iterations under Eq.~\eqref{eq:adv}; each iteration samples a batch of 4 tasks from the training pool and performs 6 rollouts per task (the GRPO group, $G=6$), with learning rate $10^{-6}$, clip 0.2, and KL coefficient $10^{-3}$. Because each offline rollout contributes a single trajectory, the within-task estimands can be measured only on the online route.


\section{Experimental Setup}
\label{sec:exp}

\paragraph{Environments and models.} We use ALFWorld, TextWorld, and WebShop, each with binary episode success, and Qwen2.5-7B-Instruct \citep{Qwen2024} trained with LoRA \citep{Hu2022} at rank 32 and $\alpha=64$. We also perform experiments on a different-sized model Qwen2.5-14B-Instruct, with the same training configurations. The prompt asks for one admissible action on the first line and, on the next line, the confidence that this episode will ultimately succeed as \texttt{<conf>NN</conf>} (details in Appendix~\ref{app:hyper}).

\paragraph{Data.} Balanced training data hold ALFWorld 1,068 rows from 200 games, TextWorld 943 from 200, and WebShop 1,066 from 372. For comparison, we train the model on each environment independently for 150 steps.

\paragraph{Training ablations.} We train a format-only baseline ($w_{\text{cal}}=w_\varphi=0$) to compare against four calibration experiments: Brier+$\varphi$ ($w_{\text{cal}}=1$, $w_\varphi=0.10$), balanced Brier, Brier+dispersion, and balanced+dispersion. Every calibration experiment carries a spread or dispersion term at $w_\varphi=0.10$, so there is no pure-Brier online arm, and each online arm uses a single training seed. Three things were fixed in advance: (i) the estimand (within-task AUROC at $k\le2$), (ii) the checkpoint rule for the balanced, dispersion, and balanced+dispersion arms (validation argmax over checkpoints $\{40,80,120\}$), and (iii) the evaluation criterion.

\paragraph{Baselines.} We compare against untrained and a format reward-only trained model variants. We also compare against two white-box signals available offline: negative mean token entropy and sequence log-probability. Finally, we compare against a depth-only model-free baseline and a depth + length texture regression, which capture what episode position \& output length alone predict.


\begin{table}[t]
\centering\small
\setlength{\tabcolsep}{5pt}
\begin{tabular}{lcccc}
\toprule
 & \multicolumn{2}{c}{Pooled AUROC} & \multicolumn{2}{c}{Tail success lift} \\
\cmidrule(lr){2-3}\cmidrule(lr){4-5}
Environment & Untrained & Trained & Untrained & Trained \\
\midrule
ALFWorld  & 0.517 & \textbf{0.681} & 1.55\x & \textbf{4.33\x} \\
TextWorld & 0.578 & \textbf{0.722} & 1.57\x & \textbf{4.35\x} \\
WebShop   & 0.462 & \textbf{0.850} & 1.11\x & \textbf{4.35\x} \\
\bottomrule
\end{tabular}
\caption{Offline training results comparing base instruction tuned vs RL trained Qwen2.5-7B-Instruct model on held-out single-trajectory Qwen2.5-7B-Instruct recordings: pooled step-level AUROC, and the tail success lift (success rate in the top-10\% confident samples vs base success rate). Bootstrap CIs in Appendix~\ref{app:offline}.}
\label{tab:offline}
\end{table}

\begin{table}[t]
\centering\small
\setlength{\tabcolsep}{5pt}
\begin{tabular}{llcc}
\toprule
Reward & Type & AUROC \ci{95\% CI} & Format \\
\midrule
Spherical   & proper   & 0.882 \ci{.837,.921} & 1.00 \\
Brier       & proper   & 0.864 \ci{.822,.903} & 1.00 \\
Ranking     & order    & 0.845 \ci{.791,.894} & 1.00 \\
L1          & improper & 0.839 \ci{.788,.887} & 1.00 \\
Logarithmic & proper   & 0.828 \ci{.770,.880} & 1.00 \\
Decision    & --       & 0.820 \ci{.773,.863} & 1.00 \\
None (task only) & --  & 0.559 \ci{.479,.636} & 0.56 \\
Untrained   & --       & 0.502 \ci{.465,.542} & 0.08 \\
\bottomrule
\end{tabular}
\caption{Offline training results on WebShop environment across different Calibration reward forms. Pooled AUROC with 95\% step-level bootstrap intervals; ``Format" is the fraction of samples that correctly predicts the tagged confidence tokens.}
\label{tab:rewardform}
\end{table}

\begin{table}[t]
\centering\small
\setlength{\tabcolsep}{6pt}
\begin{tabular}{lcc}
\toprule
Estimator & Seed 0 & Seed 1 \\
\midrule
GRPO & 0.613 \ci{.529,.696} & 0.676 \ci{.598,.749} \\
RLOO & 0.801 \ci{.739,.856} & 0.876 \ci{.825,.922} \\
PPO, value critic & 0.500 & 0.500 \\
\bottomrule
\end{tabular}
\caption{Offline training results on WebShop environment across different advantage estimator under a matched ranking + spread reward ($w{=}0.10$). Pooled AUROC with 95\% step-level bootstrap intervals; PPO with a learned value critic fails to learn parsable response format across both seeds.}
\label{tab:estimator}
\end{table}

\section{Results}
\label{sec:results}

We first establish offline RL experiments, across all three environments, that outcome-grounded training makes verbalized confidence predictive of eventual episode success (\S\ref{sec:offline}); we then show that the effect carries over to the live agent trained online in ALFWorld, and transfers to a second live environment, TextWorld (\S\ref{sec:onpolicy}). We then analyze the trained signal (\S\ref{sec:analysis}).

\subsection{Offline RL Training}
\label{sec:offline}

Each offline training sample is a single recorded trajectory, so the within and between-task estimands of Section~\ref{sec:estimands} cannot be formed; thus we report the pooled estimand with the depth-and-length controls it demands. Outcome-grounded training makes verbalized confidence predict eventual success in every environment (Table~\ref{tab:offline}): pooled AUROC rises from 0.517, 0.578, and 0.462 to 0.681, 0.722, and 0.850, and success in the most-confident decile from 1.1--1.6\x{} to about 4.3\x{} the base rate (intervals and per-environment lifts in Appendix~\ref{app:offline}).

\textbf{Before training, a free white-box signal is competitive.} The untrained model's negated mean token entropy predicts success more accurately than its verbalized confidence in all three environments, and retains most of this power after residualizing on step depth and output length (Appendix~\ref{app:offline}). Training inverts the comparison. After calibration, the verbalized confidence exceeds the token entropy in every environment---most sharply on TextWorld (0.722 vs.\ 0.399) and WebShop (0.850 vs.\ 0.430)---because the mean content-token entropy falls roughly sixfold as the policy sharpens and the entropy predictor degrades, where the token distribution grows most confident on the steps that fail.

The pooled confidence AUROC is partly explained by trajectory depth. A predictor built from step depth and output length alone, with no access to the confidence token, reaches AUROC 0.787, 0.855, and 0.799, matching or exceeding the trained confidence AUROC on ALFWorld and TextWorld though not on WebShop (Appendix~\ref{app:offline}). A depth-independent component of the confidence nonetheless does not replicate the same result: residualizing the confidence on depth and output length leaves AUROC 0.626, 0.677, and 0.811, whereas the same residualization reduces the untrained model's confidence to 0.534, 0.479, and 0.509, near chance. The depth-independent signal is thus a product of training. Because the pooled estimand cannot separate confidence from depth, we evaluate the depth-matched and within-task estimands in the online setting, which supports them (Section~\ref{sec:onpolicy}).

The presence of a calibration term matters more than its functional form (Table~\ref{tab:rewardform}). Six reward forms---spherical, Brier, ranking, L1, logarithmic, and decision---yield overlapping AUROC between 0.820 and 0.882, statistically indistinguishable, whereas training with the task reward alone reaches only 0.559, close to the untrained 0.502. The choice of advantage estimator, by contrast, is consequential (Table~\ref{tab:estimator}): under a matched ranking-plus-spread reward, RLOO attains 0.801 and 0.876 against 0.613 and 0.676 for GRPO, and PPO with a cold-start value critic fails to learn (0.500 across two seeds) parsable response format.

\begin{table*}[t]
\centering\scriptsize
\setlength{\tabcolsep}{3pt}
\begin{tabular}{lccccccc}
\toprule
 & Episode & \multicolumn{2}{c}{Pooled AUROC, $k\le2$} & ECE & Brier & \multicolumn{2}{c}{Top-10\% success, $k\le2$} \\
\cmidrule(lr){3-4}\cmidrule(lr){7-8}
Arm & success & value & $\Delta$ & (all steps) & (all steps) & value & $\Delta$ \\
\midrule
Untrained & 0.201 & 0.689 \cit{.640,.733} & -- & 0.419 \cit{.398,.438} & 0.250 \cit{.237,.264} & 0.493 \cit{.323,.675} & -- \\
Format-only & 0.188 & 0.708 \cit{.660,.749} & $+0.019$ \cit{-.010,.051} & 0.429 \cit{.408,.447} & 0.253 \cit{.239,.267} & 0.465 \cit{.296,.626} & $-0.024$ \cit{-.094,.037} \\
Brier+$\varphi$ & 0.194 & 0.761 \cit{.716,.798} & $\mathbf{+0.072}$ \textbf{\cit{.026,.112}} & 0.115 \cit{.105,.124} & 0.052 \cit{.045,.061} & 0.561 \cit{.391,.748} & $+0.080$ \cit{-.000,.185} \\
Balanced Brier & 0.197 & 0.765 \cit{.721,.805} & $\mathbf{+0.076}$ \textbf{\cit{.038,.111}} & 0.114 \cit{.102,.125} & 0.054 \cit{.046,.064} & 0.609 \cit{.413,.765} & $\mathbf{+0.115}$ \textbf{\cit{.035,.193}} \\
Brier+dispersion & 0.189 & 0.790 \cit{.742,.828} & $\mathbf{+0.101}$ \textbf{\cit{.069,.134}} & 0.054 \cit{.050,.059} & 0.044 \cit{.034,.054} & 0.635 \cit{.426,.804} & $\mathbf{+0.159}$ \textbf{\cit{.054,.224}} \\
Balanced+dispersion & 0.203 & 0.737 \cit{.684,.783} & $\mathbf{+0.048}$ \textbf{\cit{.002,.086}} & 0.053 \cit{.049,.057} & 0.047 \cit{.037,.059} & 0.657 \cit{.448,.818} & $\mathbf{+0.167}$ \textbf{\cit{.068,.235}} \\
\bottomrule
\end{tabular}
\caption{Online training results for Qwen2.5-7B-Instruct on ALFWorld. CIs are 95\% task-level cluster bootstraps, $\Delta$ = (trained - base instruction tuned) on the shared tasks; early-window ECE, Brier, top-25\% success, and the ECE and Brier deltas are in Appendix~\ref{app:calib}.}
\label{tab:onpolicy}
\end{table*}

\begin{table*}[t]
\centering\scriptsize
\setlength{\tabcolsep}{1.8pt}
\begin{tabular}{lcccccc}
\toprule
 & Pooled & Between-task & \multicolumn{4}{c}{Within-task} \\
\cmidrule(lr){2-2}\cmidrule(lr){3-3}\cmidrule(lr){4-7}
Arm & $k{=}0$ & $\Delta$, $k\le2$ & $k\le2$ ($n$ mixed) & $\Delta$, $k\le2$ ($n$ paired) & $\Delta$, all steps & $\Delta$ depth-matched, all \\
\midrule
Untrained & 0.562 \cit{.530,.600} & -- & 0.606 (20) \cit{.515,.696} & -- & -- & -- \\
Format-only & 0.568 \cit{.534,.602} & $+0.033$ \cit{-.004,.082} & 0.571 (15) \cit{.484,.657} & $-0.054$ (8) \cit{-.118,.018} & $\mathbf{-0.077}$ \textbf{\cit{-.114,-.045}} & $+0.013$ \cit{-.080,.133} \\
Brier+$\varphi$ & 0.642 \cit{.584,.691} & $+0.037$ \cit{-.017,.085} & 0.585 (20) \cit{.483,.680} & $-0.059$ (12) \cit{-.184,.059} & $-0.013$ \cit{-.056,.029} & $+0.027$ \cit{-.059,.124} \\
Balanced Brier & 0.697 \cit{.637,.744} & $+0.032$ \cit{-.019,.077} & 0.657 (19) \cit{.582,.733} & $-0.026$ (11) \cit{-.128,.055} & $-0.038$ \cit{-.082,.005} & $+0.052$ \cit{-.034,.143} \\
Brier+dispersion & 0.700 \cit{.634,.763} & $\mathbf{+0.072}$ \textbf{\cit{.029,.115}} & 0.661 (15) \cit{.568,.757} & $-0.092$ (7) \cit{-.262,.062} & $-0.050$ \cit{-.102,.004} & $+0.035$ \cit{-.060,.140} \\
Balanced+dispersion & 0.670 \cit{.612,.722} & $+0.023$ \cit{-.038,.083} & 0.587 (20) \cit{.486,.688} & $-0.031$ (13) \cit{-.120,.058} & $\mathbf{-0.072}$ \textbf{\cit{-.114,-.031}} & $-0.026$ \cit{-.107,.062} \\
\bottomrule
\end{tabular}
\caption{Decomposition of the online training gain by estimand and step window on ALFWorld. Columns move left to right from no confound removed to difficulty and depth both removed; values are absolute AUROC except where marked $\Delta$ (trained minus untrained). Within-task columns use only tasks with both a succeeding and a failing attempt: the absolute cell averages per-task AUROC over that arm's $n$ mixed-outcome tasks ($n$ mixed), and each $\Delta$ resamples the tasks mixed in \emph{both} the trained and untrained arm ($n$ paired); depth-matched is concordance over same-task, same-step-index pairs across all steps. Bold = paired 95\% CI excludes 0.}
\label{tab:decomp}
\end{table*}

\subsection{Online RL Training}
\label{sec:onpolicy}

\begin{figure}[t]
\centering
\includegraphics[width=\linewidth]{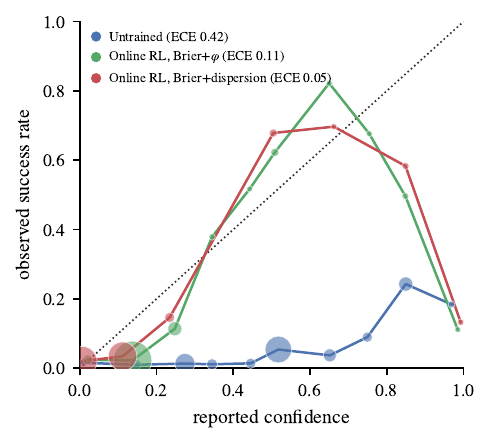}
\caption{Comparison of reliability over all steps of untrained 7B agent against two online RL trained agents on live ALFWorld. Marker area $\propto$ bin mass; same setting as of Figure~\ref{fig:riskcov}).}
\label{fig:reliability}
\end{figure}

\begin{figure}[t]
\centering
\includegraphics[width=\linewidth]{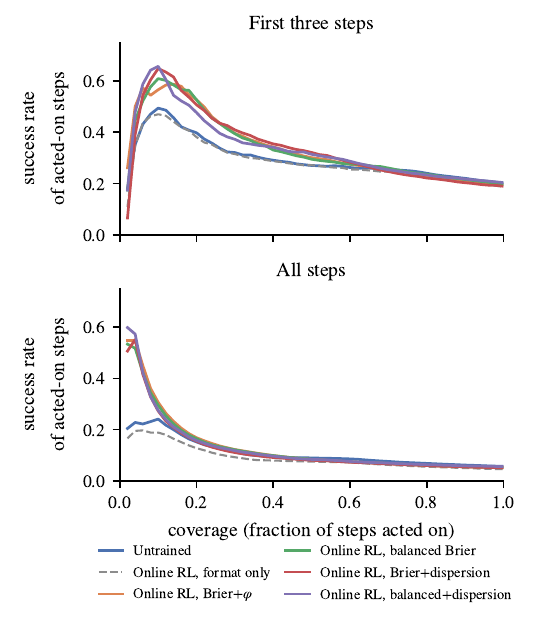}
\caption{Comparison of Risk--Coverage of different live agents on ALFWorld: for the early window ($k\le2$, top) and all steps (bottom).}
\label{fig:riskcov}
\end{figure}

\begin{figure}[t]
\centering
\includegraphics[width=0.95\linewidth]{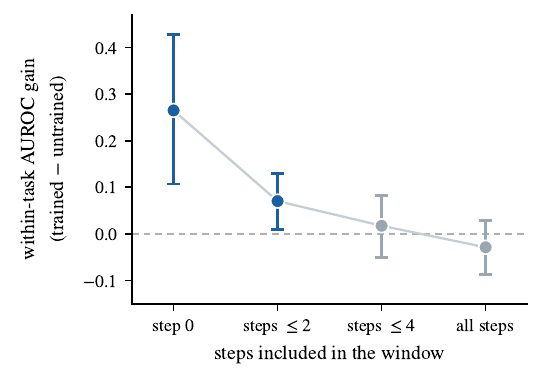}
\caption{Online training results on TextWorld environment: paired within-task AUROC gain (trained $-$ untrained) by step window; 95\% CIs over shared held-out tasks.}
\label{fig:tw_transfer}
\end{figure}

All four calibration rewards raise pooled early-window AUROC on the live acting 7B agent (Table~\ref{tab:onpolicy}): $+0.072$ \ci{.026,.112} for Brier+$\varphi$, $+0.076$ \ci{.038,.111} for balanced Brier, $+0.101$ \ci{.069,.134} for Brier+dispersion, and $+0.048$ \ci{.002,.086} for balanced+dispersion. The format-only control gives $+0.019$ \ci{-.010,.051}. Two deltas are marginal: balanced+dispersion's Pooled AUROC ($k\le2$) $+0.048$ and Brier+$\varphi$'s top-10 success ($k\le2$) \% $+0.080$ \ci{-.000,.185}.

\textbf{Calibration.} ECE over all steps (10 bins) goes from 0.419 to 0.053--0.115 and the Brier score from 0.250 to 0.044--0.054, all eight paired deltas with CIs above 0; format-only moves them by $+0.010$ and $+0.003$ respectively. Mean confidence falls from 0.475 to 0.082--0.163 against a step-level success rate of 0.057, so much of the ECE and Brier drop is the agent learning the base rate; AUROC and selective execution are the discrimination evidence (reliability diagrams in Figure~\ref{fig:reliability}). A constant prediction at the step-level base rate $p$ has ECE near zero and a Brier score of $p(1-p)$, so these two columns do not by themselves show discrimination; in the early window, where the base rate is 0.201, ECE goes from 0.277 to 0.051--0.133 and Brier from 0.229 to 0.137--0.157 (Appendix~\ref{app:calib}).

\textbf{Selective execution.} Success among the 10\% most confident early steps rises from 0.493 untrained to 0.56--0.66 across the four calibration arms, three of four deltas with CIs above 0 (Table~\ref{tab:onpolicy}, Figure~\ref{fig:riskcov}); among the top 25\% it rises from 0.363 to 0.415--0.446, all four delta CIs above 0. The format-only control moves neither window ($-0.024$ and $-0.012$), so the gain comes from the calibration term rather than the RL format reward: against that control, Brier+$\varphi$ adds $+0.053$ \ci{.016,.090} and Brier+dispersion $+0.082$ \ci{.050,.114} in pooled early-window AUROC. None of this costs task success, which holds at 0.189--0.203 against 0.201 untrained, while the \conf format parsing correctness rises from 0.978 to 1.000. The untrained agent already sorts whole tasks well: its most confident quarter succeeds at 0.557 against a 0.201 base rate. Training adds little to whole-task triage here (top-50\% 0.349--0.362 against 0.362 untrained, task-level AUROC deltas all include 0) and instead sharpens the finer step-level signal behind selective execution.

\textbf{Transfer to a second environment.} The online results above come from ALFWorld environment. To test applicability to another environment, we repeat the live, on-policy training in TextWorld with the Brier+$\varphi$ reward. Episode success rises from 0.146 to 0.262 and format correctness stays near-saturated (0.971 $\to$ 0.979). Unlike ALFWorld, where attempt-level foresight did not detectably change, within-task discrimination here improves in the early window: the paired (trained - untrained) within-task AUROC is $+0.265$ \ci{.107,.428} at step 0 and $+0.070$ \ci{.010,.130} over the first three steps, both delta CIs above 0, and decays to zero over the full episode (Figure~\ref{fig:tw_transfer}). Detailed experiment results appear in Appendix~\ref{app:tw}.

\textbf{A larger model size.} We also train the larger Qwen2.5-14B-Instruct agent online in ALFWorld with the Brier+$\varphi$ reward under the same protocol. The pooled matched-emission AUROC rises from 0.697 to 0.729 (paired trained-untrained $\Delta$ of $+0.032$), a further improvement in the same direction as the 7B gain that suggests the effect carries over to a larger model size, though fewer held-out samples make the evaluation minimal \& fluctuating; the full results are in Appendix~\ref{app:scale}.


\section{Analysis of the Trained Signal}
\label{sec:analysis}

Results in \S\ref{sec:results} show that training improves confidence. As part of this section, we ask \emph{what} it changes. We decompose the online gain by estimand to locate where the improvement comes from (\S\ref{sec:decomp}), trace it back to the first step of each episode (\S\ref{sec:step0}), and benchmark it against depth and white-box signals (\S\ref{sec:depth}) before drawing out what it means for an agent controller (\S\ref{sec:implications}).

\subsection{Decomposition}
\label{sec:decomp}

Training's gain is specifically between-task: it sharpens which tasks the agent will succeed at, not how it reads individual attempts (Table~\ref{tab:decomp}; Appendix Figure~\ref{fig:decomp}). Between-task early-window AUROC rises for every calibration arm (0.825 to 0.848--0.897), and only Brier+dispersion's $+0.072$ \ci{.029,.115} is significant. The localization suggests: the within-task estimand does not move (0.606 untrained against 0.585--0.661; paired point estimates $-0.026$ to $-0.092$), so the measured improvement is concentrated in task ranking, and no experiment met the pre-registered within-task criterion of \S\ref{sec:exp}. Consistent with this, the within-task all-step estimates match format-only experiment (format-only $-0.077$, balanced+dispersion $-0.072$) and similar to the depth-confounded estimand of \S\ref{sec:estimands}. Relative to the format-only control, the calibration term does incur gain on early-window within-task discrimination ($+0.062$ \ci{-.034,.177} for Brier+$\varphi$, $+0.127$ \ci{.062,.189} for Brier+dispersion), while the depth-matched within-task deltas are not significant (Appendix~\ref{app:onpolicy}).

Pooled and between-task early-window values move together under training (0.689/0.825 untrained to 0.761/0.862 for Brier+$\varphi$; details in Appendix~\ref{app:onpolicy}), while task-level AUROC, triage deltas, and within-task deltas show no significant gains. One reading consistent with all four estimands is that training sharpens and brings foresight forward to step 0, the task ranking the untrained agent already expresses in the early window (\S\ref{sec:step0}).

\subsection{Foresight at Step 0}
\label{sec:step0}

Training brings this task ranking forward to the very first step. Pooled step-0 AUROC rises from 0.562 \ci{.530,.600} to 0.642--0.700 across the four calibration arms, even though the confidence is emitted after the sampled first action but before any environment feedback. The signal is task-level: within-task step-0 AUROC stays at chance (0.450--0.602, every CI spanning 0.5), so the agent is ranking \emph{which tasks it will solve} rather than grading \emph{its first action}. The untrained baseline confirms this is genuine improvement: at greedy decoding it emits exactly 0.50 at step 0 on every task of a 64-task probe (Appendix~\ref{app:ladder}), so its sampled 0.562 is noise around a near-constant value, and the task ranking it expresses (between-task 0.825) emerges only after the first feedback. The mechanism follows from Eq.~\eqref{eq:adv}: uniform-outcome groups, which still arise by sampling despite mixed-outcome task selection, carry no success/failure contrast, so their gradient moves a task's confidence toward its shared outcome at every step, step 0 included.

\subsection{Depth and White-Box Baselines}
\label{sec:depth}

The depth-only baseline beats confidence on pooled all-step AUROC and matches or beats every trained experiment, yet collapses to 0.499--0.500 in the early window, whereas the early-window and depth-matched estimands are essential for a decent controller (Figure~\ref{fig:ladder}; per-size \& per-experiment values in Appendix~\ref{app:ladder},~\ref{app:onpolicy}). Token entropy is the strongest competitor for offline training; the online experiments comparison has no white-box competitor.

\subsection{Confidence as a Triage Signal}
\label{sec:implications}

Outcome-grounded RL makes verbalized confidence a calibrated early triage signal: the 10\% most confident early steps succeed at 0.56--0.66 against 0.49 untrained and a base rate of 0.20, with ECE 0.05--0.12 (Figure~\ref{fig:riskcov}). The gain is earlier recognition of which tasks the agent will fail (\S\ref{sec:step0}); attempt-level foresight does not improve on ALFWorld, though it emerges in the early window on TextWorld (\S\ref{sec:onpolicy}). Among the most confident steps (the top 10\%), higher confidence does not reliably mean higher success (Figure~\ref{fig:riskcov}), and over all steps the most-confident bin is overconfident (Figure~\ref{fig:reliability}), so a controller should rely on a band of high confidence rather than the very highest values alone. Finally, the within-task question needs training signals that contrast attempts at the same task.


\section{Conclusion}
\label{sec:conclusion}

We introduce a training method that calibrates a language agent's verbalized confidence by rewarding it against eventual task outcomes under GRPO. On a live multi-turn agent it turns an unreliable confidence token into a usable early-triage signal without hurting task success, and a decomposition shows the gain is between tasks: the agent learns earlier which tasks it will fail, not which attempts.


\section*{Limitations}

Our evidence is strongest as a between-task effect: on the live ALFWorld agent, training sharpens which tasks the agent will succeed at, and establishing the same signal at the attempt level more broadly is left to future work, as the within-task estimands move only in the early window on TextWorld and not yet on ALFWorld. Our positive live multi-turn results span two text-based environments (ALFWorld and TextWorld) at the 7B scale, and a larger 14B agent corroborates the pooled effect in Appendix~\ref{app:scale}. Our live WebShop results are inconclusive: held-out success stays low for both arms, leaving too few informative held-out tasks to assess the signal (Appendix~\ref{app:tw}); extending the live study to additional model families is left to future work.

\section*{Ethical considerations}

We conducted experiments within the provisions of the ACL Ethics Policy and relevant research-integrity guidelines. There are, to the best of our knowledge, no remaining ethical risks that have not been addressed.


\bibliography{references}

\newpage
\appendix


\section{Estimand Definitions}
\label{app:estimands}

This appendix gives the formal definitions of the four estimands and the depth-only baseline introduced in \S\ref{sec:estimands}. All five are the same AUROC statistic applied to different comparison sets and score transforms, so any difference between them is attributable to the confound each one cancels, not to a change of metric.

\paragraph{Notation.} A task $s$ is attempted $i=1,\dots,G$ times. Attempt $(s,i)$ has a binary episode outcome $Y_{s,i}\in\{0,1\}$ (success or failure) and emits a step confidence $c_{s,i,t}\in[0,1]$ at each decision step $t=0,1,\dots$; every step inherits its attempt's outcome, so step $(s,i,t)$ carries the label $Y_{s,i}$. A \emph{window} $W\subseteq\mathbb{Z}_{\ge0}$ is a set of step indices that restricts which steps are compared; we report results for three, $W_{\mathrm{all}}=\mathbb{Z}_{\ge0}=\{0,1,2,\dots\}$ (all steps), $W_{k\le2}=\{0,1,2\}$ (the early window), and $W_{0}=\{0\}$ (step~0 alone). Write $\mathcal{S}_W=\{(s,i,t): t\in W,\ c_{s,i,t}\text{ emitted}\}$ for the steps in play, and partition them by outcome into positives $\mathcal{P}_W=\{(s,i,t)\in\mathcal{S}_W : Y_{s,i}=1\}$ and negatives $\mathcal{N}_W=\{(s,i,t)\in\mathcal{S}_W : Y_{s,i}=0\}$.

\paragraph{The AUROC statistic.} Fix a set of items each carrying a real-valued score and a binary label. Thresholding the score at $\tau$ calls an item positive when its score exceeds $\tau$; sweeping $\tau$ traces the receiver operating characteristic (ROC) curve of the true-positive rate $\mathrm{TPR}(\tau)=P(\text{score}>\tau\mid\text{label}=1)$ against the false-positive rate $\mathrm{FPR}(\tau)=P(\text{score}>\tau\mid\text{label}=0)$. The area under this curve (AUROC) summarizes separation across all thresholds at once: it is $\tfrac12$ when scores are independent of the label and $1$ when every positive outscores every negative. A standard identity turns the area into a pairwise statistic rather than an integral, namely the probability that a uniformly drawn positive outranks a uniformly drawn negative \citep{Hanley1982}, which for a finite set of positive scores $A$ and negative scores $B$, with ties split evenly, is
\begin{equation}
\mathrm{AUC}(A,B)=\frac{1}{|A|\,|B|}\sum_{a\in A}\sum_{b\in B} h(a,b),
\label{eq:auc}
\end{equation}
where the pairwise kernel $h(a,b)=\mathbf{1}[a>b]+\tfrac12\mathbf{1}[a=b]$ scores a correctly ranked pair $1$, a tie $\tfrac12$, and a misranked pair $0$. Each estimand below instantiates \eqref{eq:auc} with a particular score for every step and a particular partition into $A$ and $B$.

\paragraph{Pooled AUROC.} Score each step by its raw confidence and compare every success-step against every failure-step in the window, holding none of the confounds fixed:
\begin{equation}
\mathrm{AUC}_{\mathrm{pool}}(W)=\mathrm{AUC}\big(c(\mathcal{P}_W),\,c(\mathcal{N}_W)\big).
\label{eq:pool}
\end{equation}
This conflates task difficulty~(i), attempt quality~(ii), and depth~(iii): a success-step may outrank a failure-step merely because its task is easy or its episode is nearly over.

\paragraph{Between-task AUROC.} Collapse each task to a single number by replacing every step's confidence with the mean confidence of its task over the window. Writing $\mathcal{S}_{s,W}=\{(i,t):(s,i,t)\in\mathcal{S}_W\}$ for the steps task~$s$ emits in the window,
\begin{equation}
\bar{c}_s(W)=\frac{1}{|\mathcal{S}_{s,W}|}\sum_{(i,t)\in\mathcal{S}_{s,W}} c_{s,i,t},
\label{eq:taskmean}
\end{equation}
and then apply \eqref{eq:pool} with $c_{s,i,t}$ replaced by $\bar{c}_s(W)$. Because every step of a task now shares one score, all within-task comparisons tie and contribute exactly $\tfrac12$; only cross-task comparisons carry signal, so the statistic measures whether easier (higher-mean-confidence) tasks succeed more often, i.e.\ difficulty alone.

\paragraph{Within-task AUROC.} Fix the task and compare its succeeding attempts against its failing attempts, which cancels difficulty because both sides share the task. For a task $s$ with both outcomes present in the window (a \emph{mixed-outcome} task, written $s\in\mathcal{M}_W$), define the per-task score
\begin{equation}
a_s(W)=\mathrm{AUC}\big(c(\mathcal{P}_W^{s}),\,c(\mathcal{N}_W^{s})\big),
\label{eq:within_task}
\end{equation}
and average over the mixed-outcome tasks:
\begin{equation}
\mathrm{AUC}_{\mathrm{within}}(W)=\frac{1}{|\mathcal{M}_W|}\sum_{s\in\mathcal{M}_W} a_s(W).
\label{eq:within}
\end{equation}
Only tasks that both succeed and fail in the window are estimable, so $|\mathcal{M}_W|$ (reported as the mixed count in Table~\ref{tab:ladder}) sets how many clusters the estimate rests on; it shrinks as the window narrows, which is why the step-0 and early-window within-task intervals are wide. The statistic still mixes attempt quality~(ii) with depth~(iii), since a succeeding attempt's step may outrank a failing one simply by sitting later in the episode.

\paragraph{Depth-matched within-task AUROC.} Additionally fix the step index, so only confidences emitted at the \emph{same} depth of the same task are ever compared. For each task $s$ and index $t$ with both outcomes present, let
\begin{equation}
a_{s,t}=\mathrm{AUC}\big(c(\mathcal{P}_W^{s,t}),\,c(\mathcal{N}_W^{s,t})\big),
\label{eq:depth_matched_cell}
\end{equation}
and, writing $\mathcal{C}_W=\{(s,t): t\in W,\ \exists\, i,i'\ Y_{s,i}=1,\, Y_{s,i'}=0\}$ for the qualifying (mixed-outcome, same-step) cells, average over them:
\begin{equation}
\mathrm{AUC}_{\mathrm{dm}}(W)=\frac{1}{|\mathcal{C}_W|}\sum_{(s,t)\in\mathcal{C}_W} a_{s,t}.
\label{eq:depth_matched}
\end{equation}
Neither difficulty nor depth can contribute, so what survives is attempt-level foresight alone: among attempts of one task that have reached the identical step, does the one that will ultimately succeed state higher confidence.

\paragraph{Depth-only baseline.} To quantify how much of any estimand a trivial position signal could explain, we recompute it with each confidence replaced by the negated step index,
\begin{equation}
\tilde{c}_{s,i,t}=-\,t,
\label{eq:depthonly}
\end{equation}
feeding $\tilde{c}$ in place of $c$ through the \emph{same} estimand (\eqref{eq:pool}, \eqref{eq:taskmean}, \eqref{eq:within}, or \eqref{eq:depth_matched}). This baseline ranks shallow steps above deep ones and knows nothing of the agent. Over all steps it can match or exceed a real signal, because succeeding and failing attempts differ systematically in length. In the step-0 window every $\tilde{c}=0$, so by \eqref{eq:auc} every comparison ties and the baseline is exactly $\tfrac12$; in the early window the three admissible depths leave it near chance. A trained signal that beats the depth-only baseline in the early window or at step~0 therefore exhibits foresight that position cannot explain.


\section{Untrained Estimand Ladder}
\label{app:ladder}

\begin{table*}[t]
\centering\footnotesize
\setlength{\tabcolsep}{3pt}
\begin{tabular}{lcc}
\toprule
Estimand & 7B (128 tasks) & 14B (128 tasks) \\
\midrule
Pooled, all steps & 0.740 \ci{.704,.776} & 0.698 \ci{.650,.746} \\
Depth-only baseline, all steps & 0.817 \ci{.777,.857} & 0.755 \ci{.722,.792} \\
Pooled, $k\le2$ & 0.689 \ci{.640,.733} & 0.722 \ci{.667,.774} \\
Depth-only baseline, $k\le2$ & 0.499 & 0.500 \\
Pooled, step 0 & 0.562 \ci{.530,.600} & 0.599 \ci{.532,.677} \\
Between-task, $k\le2$ & 0.825 \ci{.758,.881} & 0.830 \ci{.748,.896} \\
Between-task, all steps & 0.884 \ci{.820,.937} & 0.741 \ci{.650,.824} \\
Within-task, $k\le2$ ($n$ mixed) & 0.606 \ci{.515,.696} (20) & 0.456 \ci{.395,.519} (22) \\
Within-task, all steps & 0.651 \ci{.598,.704} & 0.653 \ci{.599,.707} \\
Within-task, step 0 & 0.529 \ci{.432,.620} & 0.400 \ci{.309,.493} \\
Depth-matched within-task, all steps & 0.633 \ci{.565,.715} & 0.616 \ci{.572,.667} \\
Depth-only within-task, all steps & 0.750 & 0.683 \\
\midrule
Attempt-level, pooled & 0.798 \ci{.735,.852} & 0.787 \ci{.714,.847} \\
Attempt-level, within-task & 0.658 \ci{.509,.806} & 0.453 \ci{.335,.571} \\
\bottomrule
\end{tabular}
\caption{Untrained Qwen2.5-7B-Instruct agents on live ALFWorld: AUROC of step confidence against eventual episode success under every estimand of \S\ref{sec:estimands}.}
\label{tab:ladder}
\end{table*}

\begin{figure}[t]
\centering
\includegraphics[width=\columnwidth]{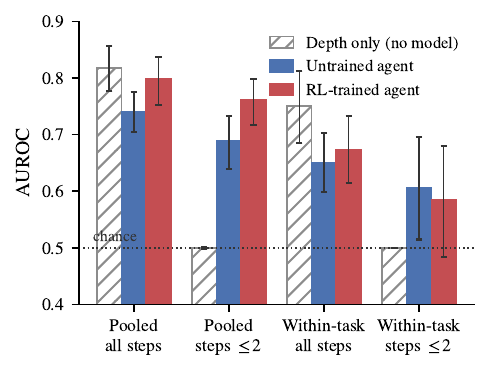}
\caption{Qwen2.5-7B-Instruct on live ALFWorld: depth-only baseline, untrained agent, and the Brier+$\varphi$ RL comparison. Bars are 95\% task-level bootstrap intervals.}
\label{fig:ladder}
\end{figure}

Table~\ref{tab:ladder} gives the untrained agents under every estimand of Section~\ref{sec:estimands}, and Figure~\ref{fig:ladder} shows the Qwen2.5-7B-Instruct agent before and after Brier+$\varphi$ training against the depth-only baseline. Untrained episode success and format correctness are 20.1\% / 97.8\% at 7B and 36.9\% / 100\% at 14B. On a 64-task probe (one run per decoding mode; no interval), greedy decoding of the untrained 7B agent gives pooled all-step AUROC 0.784 against 0.793 at temperature 1.0, and at greedy the agent emits exactly 0.50 at step 0 on every task of the probe. On the 45-task training pool the untrained 7B agent has within-task AUROC 0.677 \ci{.628,.721} at $k\le2$ and 0.763 \ci{.739,.786} over all steps.

\section{Detailed Results on Online RL Training}
\label{app:onpolicy}

\begin{table*}[t]
\centering\scriptsize
\setlength{\tabcolsep}{2pt}
\begin{tabular}{lccccccccc}
\toprule
Arm & Succ. & Pooled all & Pooled $k\le2$ & Pooled $t{=}0$ & Between $k\le2$ & Within all ($n$) & Within $k\le2$ ($n$) & Within $t{=}0$ & Depth-matched \\
\midrule
Untrained & 0.201 & 0.740 & 0.689 & 0.562 & 0.825 & 0.651 (20) & 0.606 (20) & 0.529 & 0.633 \\
 & & \cit{.704,.776} & \cit{.640,.733} & \cit{.530,.600} & \cit{.758,.881} & \cit{.598,.704} & \cit{.515,.696} & \cit{.432,.620} & \cit{.565,.715} \\[1pt]
Format-only & 0.188 & 0.754 & 0.708 & 0.568 & 0.858 & 0.665 (15) & 0.571 (15) & 0.507 & 0.646 \\
 & & \cit{.712,.792} & \cit{.660,.749} & \cit{.534,.602} & \cit{.792,.911} & \cit{.601,.727} & \cit{.484,.657} & \cit{.393,.610} & \cit{.575,.739} \\[1pt]
Brier+$\varphi$ & 0.194 & 0.799 & 0.761 & 0.642 & 0.862 & 0.673 (20) & 0.585 (20) & 0.500 & 0.659 \\
 & & \cit{.753,.838} & \cit{.716,.798} & \cit{.584,.691} & \cit{.798,.911} & \cit{.615,.733} & \cit{.483,.680} & \cit{.378,.626} & \cit{.607,.719} \\[1pt]
Balanced Brier & 0.197 & 0.782 & 0.765 & 0.697 & 0.856 & 0.683 (19) & 0.657 (19) & 0.592 & 0.685 \\
 & & \cit{.730,.829} & \cit{.721,.805} & \cit{.637,.744} & \cit{.791,.907} & \cit{.621,.753} & \cit{.582,.733} & \cit{.453,.725} & \cit{.630,.752} \\[1pt]
Brier+dispersion & 0.189 & 0.777 & 0.790 & 0.700 & 0.897 & 0.686 (15) & 0.661 (15) & 0.602 & 0.667 \\
 & & \cit{.734,.817} & \cit{.742,.828} & \cit{.634,.763} & \cit{.847,.944} & \cit{.627,.746} & \cit{.568,.757} & \cit{.483,.712} & \cit{.607,.744} \\[1pt]
Balanced+dispersion & 0.203 & 0.742 & 0.737 & 0.670 & 0.848 & 0.612 (20) & 0.587 (20) & 0.450 & 0.607 \\
 & & \cit{.696,.784} & \cit{.684,.783} & \cit{.612,.722} & \cit{.769,.911} & \cit{.564,.660} & \cit{.486,.688} & \cit{.340,.566} & \cit{.552,.665} \\
\bottomrule
\end{tabular}
\caption{Absolute values for every online RL for Qwen2.5-7B-Instruct experiment. Succ. = episode success rate; intervals are 95\% task-level cluster bootstraps}
\label{tab:app_abs}
\end{table*}

\begin{table*}[t]
\centering\scriptsize
\setlength{\tabcolsep}{3pt}
\begin{tabular}{lcccccc}
\toprule
Arm & Pooled all & Pooled $k\le2$ & Between $k\le2$ & Within all ($n$) & Within $k\le2$ ($n$) & Depth-matched \\
\midrule
Format-only & $+0.014$ & $+0.019$ & $+0.033$ & $\mathbf{-0.077}$ (8) & $-0.054$ (8) & $+0.013$ \\
 & \ci{-.019,.054} & \ci{-.010,.051} & \ci{-.004,.082} & \textbf{\ci{-.114,-.045}} & \ci{-.118,.018} & \ci{-.080,.133} \\[1pt]
Brier+$\varphi$ & $\mathbf{+0.058}$ & $\mathbf{+0.072}$ & $+0.037$ & $-0.013$ (12) & $-0.059$ (12) & $+0.027$ \\
 & \textbf{\ci{.014,.103}} & \textbf{\ci{.026,.112}} & \ci{-.017,.085} & \ci{-.056,.029} & \ci{-.184,.059} & \ci{-.059,.124} \\[1pt]
Balanced Brier & $+0.041$ & $\mathbf{+0.076}$ & $+0.032$ & $-0.038$ (11) & $-0.026$ (11) & $+0.052$ \\
 & \ci{-.008,.084} & \textbf{\ci{.038,.111}} & \ci{-.019,.077} & \ci{-.082,.005} & \ci{-.128,.055} & \ci{-.034,.143} \\[1pt]
Brier+dispersion & $+0.037$ & $\mathbf{+0.101}$ & $\mathbf{+0.072}$ & $-0.050$ (7) & $-0.092$ (7) & $+0.035$ \\
 & \ci{-.007,.084} & \textbf{\ci{.069,.134}} & \textbf{\ci{.029,.115}} & \ci{-.102,.004} & \ci{-.262,.062} & \ci{-.060,.140} \\[1pt]
Balanced+dispersion & $+0.002$ & $\mathbf{+0.048}$ & $+0.023$ & $\mathbf{-0.072}$ (13) & $-0.031$ (13) & $-0.026$ \\
 & \ci{-.044,.047} & \textbf{\ci{.002,.086}} & \ci{-.038,.083} & \textbf{\ci{-.114,-.031}} & \ci{-.120,.058} & \ci{-.107,.062} \\
\bottomrule
\end{tabular}
\caption{Paired deltas (trained - untrained) for every online RL Qwen2.5-7B-Instruct experiment; same setting as Table~\ref{tab:app_abs}. Pooled and between-task deltas resample the shared evaluation tasks; within-task deltas resample the $n$ tasks mixed in both arms.}
\label{tab:app_delta}
\end{table*}

\begin{figure}[t]
\centering
\includegraphics[width=\columnwidth]{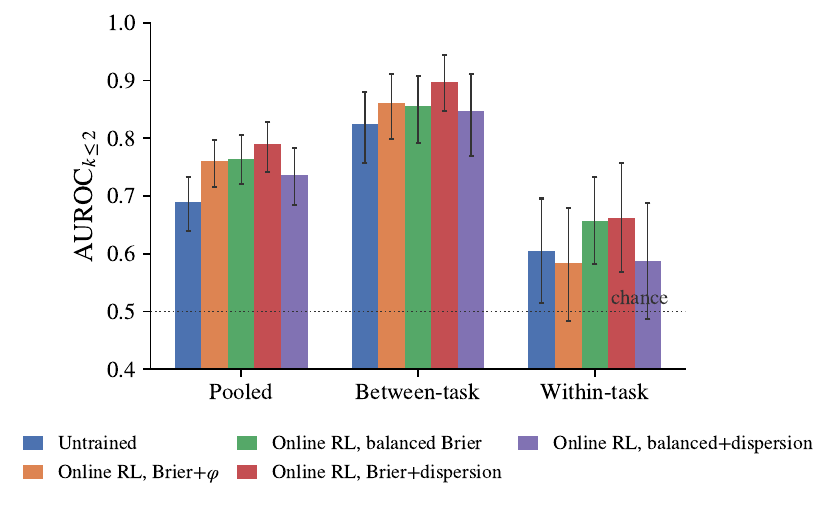}
\caption{Pooled, between-task, and within-task AUROC in the early window ($k\le2$) for the untrained Qwen2.5-7B-Instruct agent and four calibration experiments (same setting as Table~\ref{tab:onpolicy}). Bars are 95\% task-level bootstrap intervals.}
\label{fig:decomp}
\end{figure}

\begin{figure}[t]
\centering
\includegraphics[width=\columnwidth]{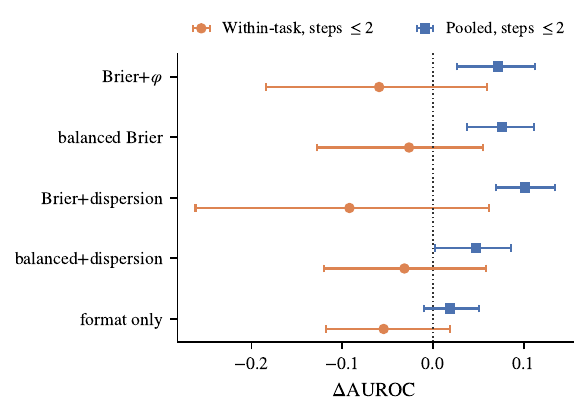}
\caption{Paired deltas of AUROC (trained - untrained) in the early window ($k\le2$), pooled (squares) and within-task (circles), with 95\% task-level bootstrap intervals.}
\label{fig:deltas}
\end{figure}

Table~\ref{tab:app_abs} gives absolute values for all six experiments under all estimands, and Table~\ref{tab:app_delta} the paired deltas; Figure~\ref{fig:decomp} plots the early-window absolute values and Figure~\ref{fig:deltas} the pooled and within-task early-window deltas. Format correctness is 0.978 untrained and 1.000 for every trained experiment, and the depth-only baseline on each experiment's rows gives pooled all-step AUROC 0.817 / 0.853 / 0.844 / 0.845 / 0.827 / 0.833 in row order and 0.499--0.500 at $k\le2$. Its within-task all-step trajectory is non-monotone (untrained 0.651 $\to$ checkpoint 40 0.677 $\to$ 80 0.745 $\to$ 120 0.673), as is its within-task early-window trajectory (0.606 $\to$ 0.603 $\to$ 0.649 $\to$ 0.585). Training dynamics for Brier+$\varphi$ (one training seed; no interval): mean confidence 0.71 at iteration 0, 0.54 at 28, 0.29 at 60, 0.20 at 86, and about 0.15 at the end; KL to the reference policy 0.002 $\to$ 0.32. Attempt-level scores (mean confidence over steps $\le2$ per attempt, one score per attempt) agree with the step-level estimands: pooled deltas are $+0.051$ \ci{-.004,.105}, $+0.099$ \ci{.057,.141}, and $+0.066$ \ci{.017,.111} for Brier+$\varphi$, Brier+dispersion, and balanced Brier, and within-task deltas are $-0.163$ \ci{-.376,.054}, $-0.246$ \ci{-.489,-.021} (7 tasks), and $-0.036$ \ci{-.205,.118}. Calibration-reward-minus-format-only contrasts: Brier+$\varphi$ minus format, within all $+0.068$ \ci{.025,.107} (8), within $k\le2$ $+0.062$ \ci{-.034,.177} (8), pooled $k\le2$ $+0.053$ \ci{.016,.090}; Brier+dispersion minus format, within $k\le2$ $+0.127$ \ci{.062,.189} (5), within all $+0.036$ \ci{.005,.068} (5), pooled $k\le2$ $+0.082$ \ci{.050,.114}.

\section{Detailed results on Other Environments}
\label{app:tw}

Section~\ref{sec:onpolicy} reports the live, on-policy TextWorld transfer in brief. Table~\ref{tab:tw_online} gives the win and emission rates and the within-task AUROC by window. Beyond the early-window gain of Figure~\ref{fig:tw_transfer}, the paired trained-minus-untrained within-task AUROC is $+0.017$ \ci{-.050,.082} by five steps and $-0.029$ \ci{-.087,.029} over all steps, and the all-step within-task AUROC is unchanged (0.697 $\to$ 0.668), so discrimination is concentrated in the early window, as in ALFWorld.

\begin{table}[t]
\centering\scriptsize
\setlength{\tabcolsep}{3pt}
\begin{tabular}{lcccc}
\toprule
 & Win & Emit & \multicolumn{2}{c}{Within-task AUROC} \\
\cmidrule(lr){4-5}
Arm & rate & rate & $k\le2$ & all steps \\
\midrule
Untrained & 0.146 & 0.971 & 0.476 \cit{.433,.526} & 0.697 \cit{.668,.727} \\
Brier+$\varphi$ & \textbf{0.262} & 0.979 & 0.506 \cit{.464,.548} & 0.668 \cit{.633,.703} \\
Format-only & 0.172 & 0.744 & 0.517 \cit{.488,.545} & 0.684 \cit{.647,.720} \\
\bottomrule
\end{tabular}
\caption{Online RL training results on TextWorld: win and emission rates over held-out tasks, and within-task AUROC by window. The paired early-window gain is in Figure~\ref{fig:tw_transfer}.}
\label{tab:tw_online}
\end{table}

The early-window improvement might instead reflect de-censoring, where training raises the fraction of early steps that emit a confidence tag, rather than the calibration term in isolation. We probe this with a format-only control but cannot cleanly separate the two. Pooled over all emitting steps the three-way comparison is null (base 0.723, full 0.702; flip full$-$base $-0.021$ \ci{-.052,.011}), so the pooled signal is not a censoring artifact. For the early-window gain we evaluate the control at an emission-matched checkpoint (tag rate 0.967, within 0.012 of both the base and full arms): neither the de-censoring component (format $-$ base, $+0.006$ \ci{-.020,.034}) nor the calibration component (full $-$ format, $-0.027$ \ci{-.055,.000}) is individually resolvable. By the trained endpoint the control's own format correctness rate falls to 0.744, which confounds a like-for-like comparison. We therefore report the early-window transfer gain (within-task AUROC $+0.265$ \ci{.107,.428} at step 0 and $+0.070$ \ci{.010,.130} over steps 0--2) without attributing it to de-censoring or calibration in isolation.

\paragraph{WebShop.} On a live, on-policy WebShop run with the binary \texttt{won} reward, held-out episode success stays near the floor for both arms (0.9\% untrained, 1.6\% Brier+$\varphi$), leaving too little success variation to estimate a within-task early-window gain: only two to three held-out tasks contain both a successful and a failed seed, so the within-task AUROC is not meaningfully identified. Step-level pooled discrimination is 0.825 \ci{.785,.889} untrained and 0.877 \ci{.784,.951} trained, with overlapping intervals due to insufficient held-out samples.


\section{Across Model Size}
\label{app:scale}

We repeat the online ALFWorld comparison on Qwen2.5-14B-Instruct, evaluating the untrained agent and a Brier+$\varphi$ checkpoint under the same held-out protocol as the 7B experiments. Episode success is 36.9\% untrained and 38.1\% trained, with format parsing rate of 1.000 in both experiments. The pooled matched-emission AUROC is 0.697 \ci{.615,.781} untrained and 0.729 \ci{.635,.818} trained; the paired (trained-untrained) $\Delta$ over the shared held-out tasks is $+0.032$ \ci{-.016,.073}, directionally consistent with the 7B pooled gain. The within-task early-window estimand is not identifiable here, as only six untrained and seven trained held-out tasks contain both a successful and a failed seed, too few for a reliable paired within-task delta. At 14B the trained agent still beats its untrained counterpart on both episode success and pooled AUROC, in the same direction as the 7B gain, so the pooled effect does not vanish with scale. Resolving the within-task estimand at this scale would require more held-out tasks across additional sizes and model families.


\section{Calibration and Selective Execution}
\label{app:calib}

\begin{table*}[t]
\centering\scriptsize
\setlength{\tabcolsep}{2pt}
\begin{tabular}{lccccccccc}
\toprule
Arm & Mean conf & ECE all & ECE $k\le2$ & Brier all & Brier $k\le2$ & Top-10\% $k\le2$ & Top-25\% $k\le2$ & Top-10\% all & Top-5\% all \\
 & all / $k\le2$ & & & & & & & & \\
\midrule
Untrained & 0.475 / 0.477 & 0.419 & 0.277 & 0.250 & 0.229 & 0.493 & 0.363 & 0.241 & 0.241 \\
 & & \cit{.398,.438} & \cit{.217,.330} & \cit{.237,.264} & \cit{.212,.248} & \cit{.323,.675} & \cit{.241,.475} & \cit{.151,.327} & \cit{.151,.333} \\[1pt]
Format-only & 0.474 / 0.476 & 0.429 & 0.290 & 0.253 & 0.229 & 0.465 & 0.363 & 0.193 & 0.182 \\
 & & \cit{.408,.447} & \cit{.236,.347} & \cit{.239,.267} & \cit{.212,.248} & \cit{.296,.626} & \cit{.220,.464} & \cit{.117,.272} & \cit{.114,.275} \\[1pt]
Brier+$\varphi$ & 0.161 / 0.189 & 0.115 & 0.051 & 0.052 & 0.137 & 0.561 & 0.444 & 0.303 & 0.491 \\
 & & \cit{.105,.124} & \cit{.041,.084} & \cit{.045,.061} & \cit{.107,.168} & \cit{.391,.748} & \cit{.299,.599} & \cit{.192,.428} & \cit{.326,.655} \\[1pt]
Balanced Brier & 0.163 / 0.183 & 0.114 & 0.054 & 0.054 & 0.139 & 0.609 & 0.444 & 0.290 & 0.469 \\
 & & \cit{.102,.125} & \cit{.042,.097} & \cit{.046,.064} & \cit{.108,.173} & \cit{.413,.765} & \cit{.293,.590} & \cit{.182,.417} & \cit{.303,.626} \\[1pt]
Brier+dispersion & 0.085 / 0.110 & 0.054 & 0.116 & 0.044 & 0.140 & 0.635 & 0.446 & 0.272 & 0.480 \\
 & & \cit{.050,.059} & \cit{.066,.164} & \cit{.034,.054} & \cit{.101,.178} & \cit{.426,.804} & \cit{.307,.585} & \cit{.167,.395} & \cit{.295,.652} \\[1pt]
Balanced+dispersion & 0.082 / 0.100 & 0.053 & 0.133 & 0.047 & 0.157 & 0.657 & 0.415 & 0.273 & 0.490 \\
 & & \cit{.049,.057} & \cit{.085,.186} & \cit{.037,.059} & \cit{.118,.199} & \cit{.448,.818} & \cit{.281,.557} & \cit{.173,.398} & \cit{.308,.673} \\
\bottomrule
\end{tabular}
\caption{Calibration and selective execution for every online training on Qwen2.5-7B-Instruct experiment; same setting as Table~\ref{tab:app_abs}. ECE uses 10 equal-width bins, top-$q$\% is the success rate among the $q$\% most confident steps in the window, and intervals are 95\% task-level cluster bootstraps.}
\label{tab:app_calib}
\end{table*}

\begin{table*}[t]
\centering\scriptsize
\setlength{\tabcolsep}{2pt}
\begin{tabular}{lccccccccc}
\toprule
Arm & ECE all & ECE $k\le2$ & Brier all & Brier $k\le2$ & Top-10\% $k\le2$ & Top-25\% $k\le2$ & Top-10\% all & Triage top-50\% & Task AUROC \\
\midrule
Format-only & $+0.010$ & $+0.013$ & $+0.003$ & $-0.001$ & $-0.024$ & $-0.012$ & $\mathbf{-0.047}$ & $-0.008$ & $+0.076$ \\
 & \cit{-.001,.021} & \cit{-.005,.030} & \cit{-.005,.010} & \cit{-.008,.006} & \cit{-.094,.037} & \cit{-.050,.036} & \textbf{\cit{-.084,-.012}} & \cit{-.034,.018} & \cit{-.013,.176} \\[1pt]
Brier+$\varphi$ & $\mathbf{-0.304}$ & $\mathbf{-0.226}$ & $\mathbf{-0.198}$ & $\mathbf{-0.093}$ & $+0.080$ & $\mathbf{+0.081}$ & $\mathbf{+0.061}$ & $-0.010$ & $-0.008$ \\
 & \textbf{\cit{-.318,-.286}} & \textbf{\cit{-.271,-.147}} & \textbf{\cit{-.213,-.180}} & \textbf{\cit{-.126,-.057}} & \cit{-.000,.185} & \textbf{\cit{.038,.160}} & \textbf{\cit{.018,.128}} & \cit{-.026,.008} & \cit{-.127,.111} \\[1pt]
Balanced Brier & $\mathbf{-0.305}$ & $\mathbf{-0.224}$ & $\mathbf{-0.196}$ & $\mathbf{-0.090}$ & $\mathbf{+0.115}$ & $\mathbf{+0.076}$ & $\mathbf{+0.050}$ & $-0.013$ & $+0.015$ \\
 & \textbf{\cit{-.319,-.288}} & \textbf{\cit{-.273,-.126}} & \textbf{\cit{-.213,-.180}} & \textbf{\cit{-.125,-.054}} & \textbf{\cit{.035,.193}} & \textbf{\cit{.042,.155}} & \textbf{\cit{.007,.111}} & \cit{-.034,.010} & \cit{-.092,.124} \\[1pt]
Brier+dispersion & $\mathbf{-0.364}$ & $\mathbf{-0.162}$ & $\mathbf{-0.206}$ & $\mathbf{-0.089}$ & $\mathbf{+0.159}$ & $\mathbf{+0.092}$ & $+0.035$ & $-0.008$ & $+0.098$ \\
 & \textbf{\cit{-.382,-.346}} & \textbf{\cit{-.270,-.057}} & \textbf{\cit{-.225,-.187}} & \textbf{\cit{-.131,-.049}} & \textbf{\cit{.054,.224}} & \textbf{\cit{.041,.143}} & \cit{-.012,.095} & \cit{-.031,.016} & \cit{-.006,.210} \\[1pt]
Balanced+dispersion & $\mathbf{-0.366}$ & $\mathbf{-0.144}$ & $\mathbf{-0.203}$ & $\mathbf{-0.072}$ & $\mathbf{+0.167}$ & $\mathbf{+0.095}$ & $+0.033$ & $0.000$ & $+0.047$ \\
 & \textbf{\cit{-.384,-.346}} & \textbf{\cit{-.239,-.033}} & \textbf{\cit{-.221,-.184}} & \textbf{\cit{-.114,-.025}} & \textbf{\cit{.068,.235}} & \textbf{\cit{.019,.119}} & \cit{-.013,.105} & \cit{-.023,.018} & \cit{-.064,.155} \\
\bottomrule
\end{tabular}
\caption{Paired deltas (trained - untrained) of the calibration and selective-execution metrics; same setting as Table~\ref{tab:app_abs}. Task AUROC = AUROC of mean early confidence against ``task solved at least once''.}
\label{tab:app_calib_delta}
\end{table*}

\begin{table*}[t]
\centering\footnotesize
\setlength{\tabcolsep}{6pt}
\begin{tabular}{lcc}
\toprule
 & Qwen2.5-7B-Instruct & Qwen2.5-14B-Instruct \\
\midrule
Mean conf, all / $k\le2$ & 0.475 / 0.477 & 0.534 / 0.646 \\
Step success, all / $k\le2$ & 0.057 / 0.201 & 0.112 / 0.289 \\
ECE, all & 0.419 \ci{.398,.438} & 0.425 \ci{.388,.458} \\
ECE, $k\le2$ & 0.277 \ci{.217,.330} & 0.357 \ci{.296,.417} \\
Brier, all & 0.250 \ci{.237,.264} & 0.303 \ci{.281,.328} \\
Brier, $k\le2$ & 0.229 \ci{.212,.248} & 0.310 \ci{.283,.336} \\
Top-10\% succ., $k\le2$ & 0.493 \ci{.323,.675} & 0.717 \ci{.556,.839} \\
Top-25\% succ., $k\le2$ & 0.363 \ci{.241,.475} & 0.571 \ci{.406,.724} \\
Top-10\% succ., all & 0.241 \ci{.151,.327} & 0.275 \ci{.192,.375} \\
Triage, top-50\% tasks & 0.362 \ci{.245,.497} & 0.464 \ci{.328,.591} \\
Triage, top-25\% tasks & 0.557 \ci{.375,.714} & 0.755 \ci{.521,.964} \\
Task-level AUROC & 0.680 \ci{.578,.787} & 0.756 \ci{.663,.842} \\
Tasks solved at least once & 38/128 & 51/128 \\
\bottomrule
\end{tabular}
\caption{Calibration and selective execution of the untrained Qwen2.5-7B-Instruct and Qwen2.5-14B-Instruct agents on the held-out set. Intervals are 95\% task-level cluster bootstraps.}
\label{tab:app_untrained_calib}
\end{table*}

Tables~\ref{tab:app_calib} and~\ref{tab:app_calib_delta} give the calibration and selective-execution metrics for all experiments with paired deltas, and Table~\ref{tab:app_untrained_calib} gives the untrained block at two model sizes. Full-coverage step-level success is 0.046--0.057 over all steps and 0.186--0.203 at $k\le2$, and task-level triage (success if only the top 50\% of tasks by mean early confidence are attempted) is 0.362 / 0.354 / 0.352 / 0.349 / 0.354 / 0.362 in row order. The task-level triage deltas (format-only $-0.008$, calibration arms $-0.013$ to $0.000$) and the task-level AUROC deltas ($-0.008$ to $+0.098$) all have CIs including 0.


\section{Details on Offline RL Training}
\label{app:offline}

\paragraph{Pooled AUROC intervals.} The 95\% step-level bootstrap intervals for Table~\ref{tab:offline}, untrained vs.\ trained, are: ALFWorld 0.517 \ci{.496,.538} vs.\ 0.681 \ci{.662,.699} at step 300 (0.689 \ci{.671,.706} at step 150); TextWorld 0.578 \ci{.547,.611} vs.\ 0.722 \ci{.692,.753}; WebShop 0.462 \ci{.396,.550} vs.\ 0.850 \ci{.773,.910} (18--23 successes, hence the wide interval). Untrained-to-trained top-10\% tail success lift is 1.55\x$\to$4.33\x{} (ALFWorld; 2.73\x{} at step 150), 1.57\x$\to$4.35\x{} (TextWorld), and 1.11\x$\to$4.35\x{} (WebShop).

\paragraph{Additional signals on the untrained rows.} Negative mean token entropy gives AUROC 0.741 \ci{.722,.762} (ALFWorld), 0.667 \ci{.630,.703} (TextWorld), and 0.806 \ci{.724,.881} (WebShop), with top-10\% multipliers 4.22\x, 2.97\x, and 3.34\x. Sequence log-probability gives AUROC 0.706 \ci{.685,.726} (ALFWorld), 0.593 \ci{.560,.627} (TextWorld), and 0.762 \ci{.653,.858} (WebShop), with top-10\% multipliers 3.10\x, 1.91\x, and 3.34\x. Negative last-token entropy gives 0.531, 0.493, and 0.401, with multipliers 0.81\x, 1.06\x, and 0.0\x. Untrained confidence on emitted steps only gives 0.610 \ci{.579,.639} and 1.93\x{} in ALFWorld (6,184 steps / 410 successes) and 0.619 \ci{.586,.652} and 1.61\x{} in TextWorld (2,383 / 180); the WebShop emitted subset (265 steps, 2 successes) is not estimable.

\paragraph{Texture control.} The AUROC of the OLS prediction of success from step depth and $\log(1+\text{content tokens})$ alone is 0.769 / 0.849 / 0.858 on the untrained rows and 0.787 / 0.855 / 0.799 on the trained rows (ALFWorld / TextWorld / WebShop). Residualized untrained confidence gives 0.534 \ci{.510,.557} / 0.479 \ci{.440,.516} / 0.509 \ci{.386,.632}; residualized untrained token entropy 0.707 \ci{.687,.727} / 0.586 \ci{.549,.623} / 0.810 \ci{.729,.882}; residualized trained confidence 0.626 \ci{.603,.649} / 0.677 \ci{.638,.715} / 0.811 \ci{.712,.887}.

\paragraph{White-box signals on the trained rows.} On the trained models' rows, ALFWorld (step 150) token entropy gives 0.660 and sequence log-probability 0.654 against trained confidence 0.689; TextWorld entropy 0.399 against 0.722; WebShop entropy 0.430 against 0.850. Mean content-token entropy on ALFWorld falls from 0.145 to 0.025.

\paragraph{ALFWorld checkpoint curve.} On the same held-out rows, pooled AUROC raw $\to$ residualized is 0.517 $\to$ 0.534 untrained; 0.690 $\to$ 0.634 at step 100; 0.689 $\to$ 0.626 at 150; 0.659 $\to$ 0.534 at 200; 0.677 $\to$ 0.612 at 250; 0.681 $\to$ 0.633 at 300. Success among the top 5\% of steps goes from 0.137 untrained to 0.310 at step 300, and among the top 20\% from 0.109 to 0.181; at step 150, success among the top 25\% is 0.165 against 0.096 untrained.

\paragraph{Separate held-out probes.} A second set of WebShop and TextWorld probes uses a different held-out split and counts WebShop wins differently (96 of 1,303 steps), so its values are not comparable to Table~\ref{tab:offline}. On that split: WebShop untrained AUROC 0.518 \ci{.485,.554} with ECE 0.452; ranking reward at 150 steps 0.728 \ci{.676,.778}, ECE 0.185, top-10\% success 0.162 (2.2\x{} of 0.074); ranking + spread $w=0.10$ at 150 steps 0.795 \ci{.738,.844}, ECE 0.146, top-10\% 0.369 (5.0\x). TextWorld untrained 0.522 \ci{.438,.610} with ECE 0.586; ranking at 150 steps 0.775 \ci{.690,.853}, ECE 0.100, top-10\% 0.333 (5.4\x{} of 0.062), on 600 steps with 37 successes.


\section{Ablations}
\label{app:spread}

\paragraph{Learning rate and probe temperature.} The Brier arm's held-out WebShop AUROC rises steeply near a learning rate of $2\times10^{-6}$ and plateaus by $5\times10^{-6}$ at about 0.86 (one seed; no interval). Reading its confidence greedily rather than at $T=0.8$ barely moves it: 0.863 \ci{.819,.901} greedy against 0.846 \ci{.795,.892} at $T=0.8$ (emitted steps, $n=217$ and 219), and 0.864 under the pooled convention of Table~\ref{tab:rewardform} (missing tag scored 0.5) for the same checkpoint.

\begin{table}[t]
\centering\scriptsize
\setlength{\tabcolsep}{2.5pt}
\begin{tabular}{ccccc}
\toprule
 & AUROC, sampled & Mass in & Mass & AUROC, greedy \\
$w$ & (per seed) & $[0.2,0.8]$ & $\ge0.8$ & (2 seeds) \\
\midrule
0.00 & 0.746 (0.717, 0.775) & 0.64 & 0.34 & 0.720 / 0.762 \\
0.05 & 0.678 (0.679, 0.676) & 0.76 & 0.23 & 0.690 / 0.698 \\
0.10 & 0.675 (0.675)        & 0.74 & 0.25 & 0.663 / 0.676 \\
0.15 & 0.584 (0.594, 0.574) & 0.88 & 0.12 & 0.593 / 0.597 \\
0.30 & 0.556 (0.587, 0.525) & 0.93 & 0.08 & 0.568 / 0.520 \\
\bottomrule
\end{tabular}
\caption{Spread-weight sweep, WebShop offline RL training, ranking reward, evaluated on the held-out prefixes of Table~\ref{tab:rewardform}; sampled columns use $K=8$ decodes per prefix at temperature $T=1.0$, top-$p$ 0.95, seed-averaged ($w=0.10$ has one sampled seed), and the interior mass column is the closed interval $[0.2,0.8]$ (so $c=0.8$ is counted in both mass columns).}
\label{tab:spread}
\end{table}

Table~\ref{tab:spread} gives the spread-weight sweep. The second greedy $w=0.10$ value is the run reported as GRPO seed 1 in Table~\ref{tab:estimator}, and a third run of that configuration gives 0.613 (GRPO seed 0 there); the same sweep on ALFWorld offline (checkpoint 150, 176 held-out prefixes / 58 successes, 3 seeds, 15 cells) gives 0.760--0.822 with overlapping intervals. Reliability by bin for that sweep ($K=8$ sampled decodes at $T=1.0$, emitted decodes only). At $w=0$: confidences $\ge0.8$ hold 0.39 of the mass with mean 0.86 and success 0.33; $0.2\le c<0.8$ holds 0.60 with mean 0.61 and success 0.20; $c<0.2$ holds 0.01. At $w=0.10$: 0.29 / 0.86 / 0.37; 0.70 / 0.59 / 0.22; 0.01. The untrained model at $T=1.0$ emits a tag on 17\% of decodes (300 of 1,760), with $c\ge0.8$ holding 0.47 of the emitted mass at success 0.31. Emission is 42\% at $w=0$ and 50\% at $w=0.10$.


\section{Hyperparameters, Prompt \& Rewards}
\label{app:hyper}

\paragraph{Prompt (used in both training \& evaluation).}
\begin{tcolorbox}[colback=gray!5,colframe=gray!50,boxrule=0.4pt,left=4pt,right=4pt,top=3pt,bottom=3pt]
\small\ttfamily
You are an agent solving a household task in a text world. Each turn you are given the task, a log of what you have already done this episode, what you currently observe, and the list of admissible actions. Choose exactly ONE admissible action that best progresses toward the task. Do not repeat actions that did not help. First output the chosen action text on its own line, copied verbatim from the admissible list, with no explanation. Then on the NEXT line output your confidence that THIS EPISODE will ultimately succeed, as an integer 0-100 wrapped exactly as \texttt{<conf>NN</conf>}. Output nothing else.
\end{tcolorbox}

\paragraph{Models.} Qwen2.5-7B-Instruct and Qwen2.5-14B-Instruct; LoRA rank 32, $\alpha=64$ for all training.

\paragraph{Online training.} 120 iterations $\times$ 4 tasks $\times$ 6 attempts (the group); advantage per step as in Eq.~\eqref{eq:adv}; reward as in Eq.~\eqref{eq:reward}; learning rate $10^{-6}$; PPO-style clip 0.2; KL-to-reference coefficient $10^{-3}$; gradient-norm clip 1.0; micro-batch 1; checkpoints every 20 iterations; sampling temperature 1.0; up to 30 steps and 48 new tokens per step. Arms: format-only ($w_{\text{cal}}=0$, $w_\varphi=0$); Brier+$\varphi$ ($w_{\text{cal}}=1$, $\varphi=4c(1-c)$, $w_\varphi=0.10$); balanced Brier ($w_{\text{cal}}=1$, Brier reweighted with $0.5/\bar Y_s$ on wins and $0.5/(1-\bar Y_s)$ on losses, $\varphi=4c(1-c)$, $w_\varphi=0.10$); Brier+dispersion ($w_{\text{cal}}=1$, $\varphi$ replaced by $\min(1,4(c-\bar c_s)^2)$, $w_\varphi=0.10$); balanced+dispersion (both). Training tasks: 45 mixed-outcome tasks from seeds 18--245 (7B). Validation seeds 500--563; test seeds 1000--1127. All disjoint.

\paragraph{Offline training.} GRPO on one recorded trajectory per game from the untrained 7B policy, cut into (prefix $\to$ recorded action, outcome) rows; balanced parquets ALFWorld 1,068 rows / 200 games, TextWorld 943 / 200, WebShop 1,066 / 372 (won rate 0.35 after balancing; raw WebShop 5,191 rows / 400 games / 7.2\% won); no game has mixed outcomes. 8 generations per prefix; temperature 0.8; maximum completion 256 tokens; AdamW with a linear schedule; learning rate $5\times10^{-6}$ ($10^{-6}$ was too cold and produced artefacts); batch 8; KL coefficient $10^{-3}$; 600 steps for the WebShop ablations (checkpoint 600) and 150 for the multi-environment checkpoints. Reward = 0.5 action-match + 0.2 format + 0.3 calibration, plus $w\,\varphi(c)$ in the spread sweep; the reward-form arms of Table~\ref{tab:rewardform} use $w_{\text{cal}}=1.0$ in place of the 0.3 default. Held-out probes decode greedily (temperature 0, 64 new tokens) unless stated; the spread sweep and its reliability analysis use $K=8$ sampled decodes at $T=1.0$, top-$p$ 0.95.

\begin{sloppypar}
\paragraph{Calibration reward forms.} With $c$ the parsed confidence and $Y$ the outcome ($\epsilon$, the sigmoid gain 10, the abstain utility 0.5, and the EMA decay 0.99 are implementation constants of the reward code; $\tau=0.1$): Brier $1-(c-Y)^2$; logarithmic, normalized to $[0,1]$, $1+\big[Y\log c+(1-Y)\log(1-c)\big]/(-\log\epsilon)$ with $c$ clipped to $[\epsilon,1-\epsilon]$ and $\epsilon=10^{-3}$; spherical $q_Y/\sqrt{c^2+(1-c)^2}$ with $q_Y=c$ if $Y=1$ and $1-c$ otherwise; L1 $1-|c-Y|$; decision/selective $gY+(1-g)\cdot0.5$ with $g=\sigma\big(10\,(c-0.5)\big)$, so the optimum commits when $P(Y=1\mid h)>0.5$; ranking surrogate $\sigma\big((c-\mu_{\text{lose}})/\tau\big)$ for a winning completion and $\sigma\big((\mu_{\text{win}}-c)/\tau\big)$ for a losing one, where $\mu_{\text{win}}$ and $\mu_{\text{lose}}$ are exponential moving averages (decay 0.99) of the confidence of the opposite class and $\tau=0.1$.

\paragraph{Reward-form and estimator details (Tables~\ref{tab:rewardform} and~\ref{tab:estimator}).} Across seeds the proper forms overlap---Brier spans 0.823--0.864 and spherical 0.852--0.882, of the same order as the between-form spread---so the six forms are indistinguishable at this seed count; the paired Brier-minus-ranking gap is $+0.019$ \ci{-.024,.064} on seed 0, where the top-25\% step-success gap is $+0.018$ \ci{-.145,.073}. Task reward only gives 0.559 and 0.555 at format rates 0.56 and 0.59 over two seeds, against untrained 0.502 at 0.08. Table~\ref{tab:estimator} carries the spread term at $w=0.10$ and so is not comparable to the ranking run of Table~\ref{tab:rewardform} (0.845); the spread sweep of the same reward gives 0.663/0.676 at that weight (Appendix~\ref{app:spread}). PPO's three 0.500 runs (seed 2 at a 10\x{} value-head learning rate) reflect a cold-start value head: large, noisy advantages for roughly the first 15 updates, tag emission collapsing to a constant, and zero within-group reward variance.
\end{sloppypar}


\section{Computational Budget and Reproducibility}
\label{app:repro}

\paragraph{Models and parameters.} All training uses Qwen2.5-Instruct backbones at 7B and 14B parameters \citep{Qwen2024}, adapted with LoRA \citep{Hu2022} at rank 32 and $\alpha=64$ (Appendix~\ref{app:hyper}); only the adapter parameters are updated.

\paragraph{Hardware and compute budget.} All models were trained on a single node with 8$\times$ NVIDIA H200 GPUs. Each run is a single-GPU LoRA job (adapters only), with experiments scheduled concurrently across the eight GPUs. Representative single-GPU wall-clock is about 8\,h for a 7B online ALFWorld experiment, 12\,h for the 14B experiment, 5\,h for a TextWorld experiment, 2\,h for WebShop, and roughly 10\,min for an offline GRPO reward-form computation. By category: the offline GRPO reward-form and estimator sweeps total $\approx$2 GPU-hours; the online 7B ALFWorld arms (six reward forms plus four additional-seed runs) $\approx$79 GPU-hours; the TextWorld and WebShop online runs $\approx$18 GPU-hours; and the cross-size experiments $\approx$49 GPU-hours. Including preliminary and failed runs, total training compute is approximately 200 GPU-hours.

\paragraph{Software.} PyTorch 2.13.0 (CUDA 13.0), Transformers 5.17.0, TRL 1.14.0, PEFT 0.21.0, Accelerate 1.15.0, Datasets 5.0.1, and NumPy 2.5.3. Offline training uses TRL GRPO; the online route uses the in-house on-policy learner (Appendix~\ref{app:hyper}).

\paragraph{Runs and statistical reporting.} Each online training experiment uses a single training seed (Section~\ref{sec:exp}); the offline reward-form and estimator comparisons use the seed counts stated in Appendix~\ref{app:hyper} and Table~\ref{tab:spread}. All reported intervals are 95\% task-level cluster bootstraps over held-out tasks. Training, validation, and test seeds are disjoint (Appendix~\ref{app:hyper}); held-out evaluation seeds never appear in training.

\paragraph{Reproducibility.} Every ingredient needed to reproduce our results is specified in the paper: the on-policy learner configuration and full hyperparameters, the agent prompt template with the inline \texttt{<conf>NN</conf>} format, and the exact per-arm reward definitions (Appendix~\ref{app:hyper}); the pre-registered estimand (within-task AUROC at $k\le2$), the checkpoint-selection rule, and the disjoint train/validation/test seed ranges (Section~\ref{sec:exp}); and the 95\% task-level cluster-bootstrap procedure used for all reported intervals. We provide anonymized training and evaluation code, configurations, and prompts in the supplementary material to reproduce every experiment.

\section{Scientific Artifacts and Licenses}
\label{app:artifacts}

Our experiments build on three agentic environments and the Qwen2.5-Instruct model family \citep{Qwen2024}, adapted with LoRA \citep{Hu2022} using the TRL library \citep{vonWerra2020} for training. ALFWorld \citep{Shridhar2021}, TextWorld \citep{Cote2018}, and WebShop \citep{Yao2022} are released under the MIT License, and TRL under the Apache-2.0 License. Among the models, Qwen2.5-7B-Instruct and Qwen2.5-14B-Instruct are released under the Apache-2.0 License. We use each artifact solely for research, consistent with its intended use as an environment or model for studying language agents, and within the terms of its license. The three environments are standard, publicly available English-language task suites (household and text-adventure navigation and simulated web shopping) that carry no personally identifying information and no offensive content.

\section{Human Subjects}
\label{app:human_subjects}

Our study involves no human annotators, participants, or user-generated data.

\section{Use of AI Assistants}
\label{app:ai_assistants}

We used the AI-based coding and writing assistant Claude Code with the Claude-4.8-Opus model for routine code implementation and language editing for the paper.

\end{document}